\documentclass[11pt]{amsart}   

\usepackage{amsmath,amsfonts,bm}

\def\eqref#1{equation~\ref{#1}}

\def\1{\bm{1}}

\DeclareMathAlphabet{\mathsfit}{\encodingdefault}{\sfdefault}{m}{sl}
\SetMathAlphabet{\mathsfit}{bold}{\encodingdefault}{\sfdefault}{bx}{n}

\usepackage[T1]{fontenc}
\usepackage[utf8x]{inputenc}
\usepackage{amscd,amsfonts,amssymb,amsmath,amsthm,latexsym,mathtools}
\usepackage{times}
\usepackage{graphicx}
\usepackage{booktabs,multirow}
\usepackage{xy}
\xyoption{all}
\usepackage{url}
\usepackage{tikz}
\usepackage[small]{caption}
\usepackage{comment}
\usepackage{hyperref}

\usepackage[a4paper,top=4cm, bottom=5cm, left=2.9cm, right=2.9cm]{geometry}

\newif\ifspanish
\spanishfalse
\ifspanish
\usepackage[spanish]{babel}
\fi

\theoremstyle{plain}
\ifspanish

\else

\fi

\theoremstyle{definition}
\ifspanish

\else

\fi

\theoremstyle{remark}
\ifspanish

\else

\fi

\title{Semi-Supervised Machine Learning: a Homological Approach}
\author{
Adri\'an In\'es, C\'esar Dom\'{\i}nguez, J\'onathan Heras, Gadea Mata \and Julio Rubio
}
\date{}

\address{\small \rm  Departamento de Matemáticas y Computación. Universidad de La Rioja}
\email{\texttt{\{adrian.ines, cesar.dominguez, jonathan.heras\}@unirioja.es}}
\email{\texttt{\{gadea.mata, julio.rubio\}@unirioja.es}}
\thanks{This work was partially supported by the projects PID2020-115225RB-I00 and  PID2020-116641GB-I00, funded by MCIN/AEI/10.13039/501100011033 and by ``European Union NextGenerationEU/PRTR''}

\begin{document}

\begin{abstract}
In this paper we describe the mathematical foundations of a new approach to semi-supervised Machine Learning. Using techniques of Symbolic Computation and Computer Algebra, we apply the concept of \emph{persistent homology} to obtain a new semi-supervised learning method.
\end{abstract}

\maketitle

\section*{Introduction}

Machine Learning and Deep Learning methods have become the state-of-the-art approach for solving data classification tasks. In order to use those methods, it is necessary to acquire and label a considerable amount of data; however, this is not straightforward in some fields, since data annotation is time consuming and may require expert knowledge. This challenge can be tackled by means of semi-supervised learning methods that take advantage of both labelled and unlabelled data. In our team we have applied this Machine Learning paradigm in various applied projects (e.g. \cite{ownsemi}). In this paper, we present a new semi-supervised learning method based on techniques from Topological Data Analysis. In particular, we have used a homological approach that consists of studying the persistence diagrams associated with data from binary classification tasks using the bottleneck and Wasserstein distances. In addition, we have carried out a thorough analysis of the developed method using 5 structured datasets. The results show that the semi-supervised method developed in this work outperforms both the results obtained with models trained with only manually labelled data, and those obtained with classical semi-supervised learning methods, improving the models by up to a 16\%.

\section{Conceptual presentation}

\begin{figure}[ht]
    \centering
\includegraphics[width=12cm, height=8cm]{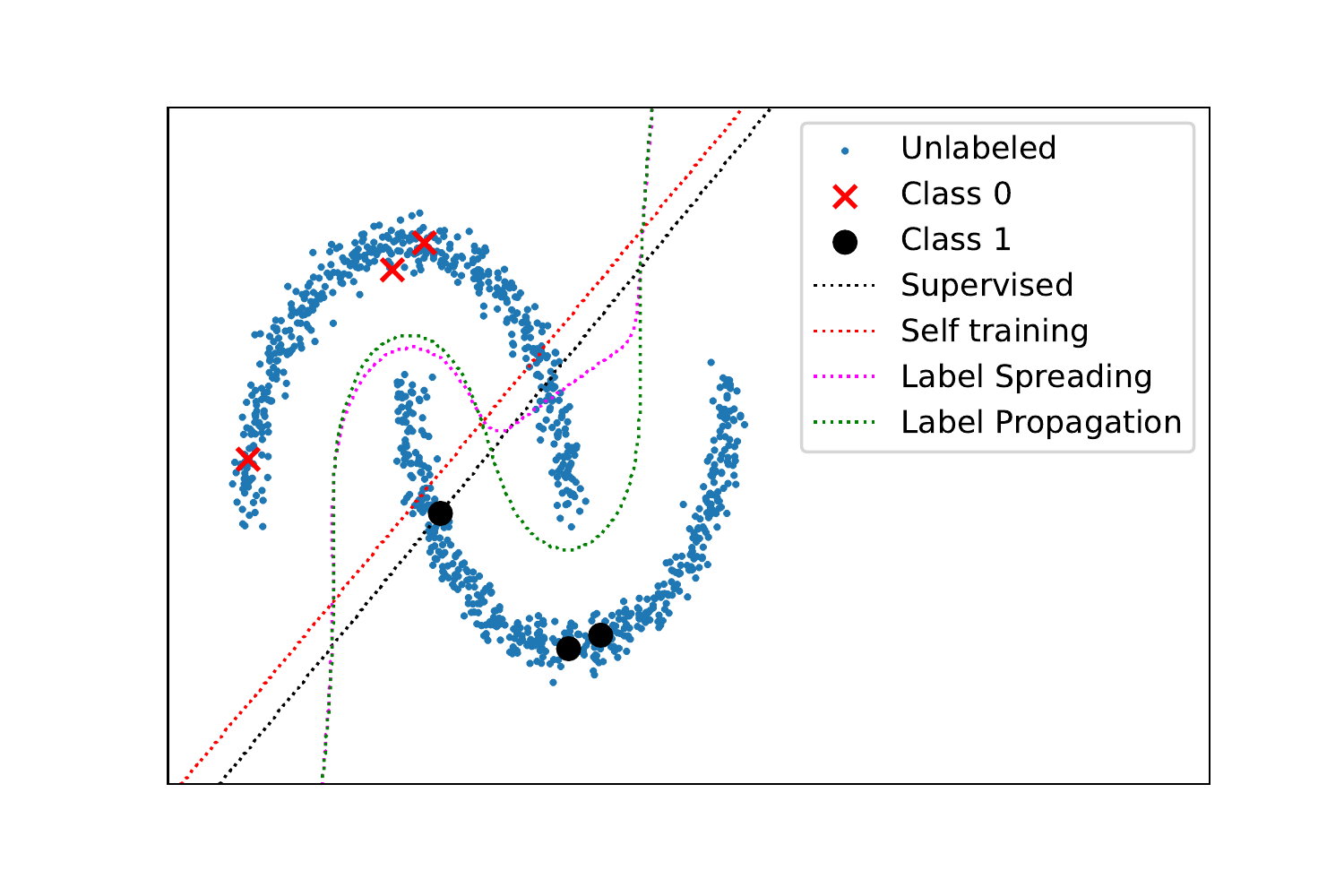}
    \caption{Example with two ``connected manifolds''}
    \label{fig:espiral}
\end{figure}

Our method falls within the discipline of Topological Data Analysis (hereinafter TDA), a field devoted to extracting topological and geometrical information from data. And the problem undertaken is motivated by the challenge of obtaining enough annotated data to apply Machine Learning techniques. To that end, a family of methods that has been successfully applied in the literature is semi-supervised learning. Semi-supervised learning methods provide a means of using unlabelled data to improve models' performance when we have access to a large corpus of data that is difficult to annotate. Traditional semi-supervised learning algorithms, such as  Label Spreading~\cite{labelSpread} and Label Propagation~\cite{labelProp}, focus on the distance among the data points to annotate unlabelled data points; i.e. on the metric and density characteristics of the data in a dataset.
However, there are contexts where metric approaches could be misleading. As shown in Figure \ref{fig:espiral}, distances are not the right discriminators in complex situations and, therefore other ideas are needed. Our inspiration comes from the \emph{Manifold Hypothesis}~\cite{fefferman2016testing}, which explores when high dimensional data could tend to lie in low dimensional manifolds. Roughly speaking, our method works under the hypothesis that each class in the dataset lies on a manifold. In particular, homological information should be respected when we add an unlabelled point to one of the classes. Our method is therefore as follows:  given two sets of data points $A$ and $B$, corresponding to the points labelled with class 1 and class 2, respectively, we  assume there are two manifolds associated with each set, ${\mathcal M}_A$ and ${\mathcal M}_B$ respectively; now, given an unlabelled data point $x$, if $x$ belongs to class 1, for instance, then $A\cup \{x\}$ would lie on a manifold more similar to ${\mathcal M}_A$ than the manifold corresponding to $B\cup \{x\}$ with respect to ${\mathcal M}_B$. 

All the code developed for this project and also the conducted experiments are available at the project webpage \url{https://github.com/adines/TTASSL}.

\section{Description of the method}

In this section, we describe the semi-supervised learning algorithm that we have designed to tackle binary classification tasks. We start with a set $X_1$ of points from class 1, a set $X_2$ of points from class 2, and a set $X$ of unlabelled points. The objective of our algorithms is to annotate the elements of $X$ by using topological properties of $X_1$ and $X_2$. We assume some familiarity with notions employed in TDA such as Vietoris-Rips filtration (we denote the Vietoris-Rips filtration associated with a set $X$ by $V_X$), persistence diagrams (we denote the persistence diagram associated with a filtration $F$ by $P(F)$), and the bottleneck and Wasserstein distances (denoted by $d_B$ and $d_W$ respectively). For a detailed introduction to these topics see~\cite{zomorodian2012topological}.

Our semi-supervised learning algorithm takes as input the sets $X_1$ and $X_2$, a point $x\in X$, a threshold value $t$, and a flag that indicates whether the bottleneck or the Wasserstein distance should be used. We denote the chosen distance as $d$. The output produced by our algorithm is whether the point $x$ belongs to $X_1$, $X_2$ or none of them. In order to decide the output of the algorithm, our hypothesis is that if a point belongs to $X_1$, analogously for $X_2$, then when adding the point to the manifold on which $X_1$ lies, the topological variation will be minimal; whereas if the point does not belong to $X_1$, the variation will be greater. In particular, we proceed as follows:

\begin{enumerate}
    \item Construct the Vietoris-Rips filtrations $V_{X_1}$, $V_{X_2}$, $V_{X_1 \cup \{x\}}$ and $V_{X_2 \cup \{x\}}$;
    \item Construct the persistence diagrams $P(V_{X_1})$, $P(V_{X_2})$, $P(V_{X_1 \cup \{x\}})$ and $P(V_{X_2 \cup \{x\}})$;
    \item Compute the distances $d(P(V_{X_1}),P(V_{X_1 \cup \{x\}}))$ and $d(P(V_{X_2}),P(V_{X_2 \cup \{x\}}))$, from now on $d_1$ and $d_2$ respectively;
    \item If both $d_1$ and $d_2$ are greater than the threshold $t$, return none; otherwise, return the set associated with the minimum of the distances $d_1$ and $d_2$.
\end{enumerate}

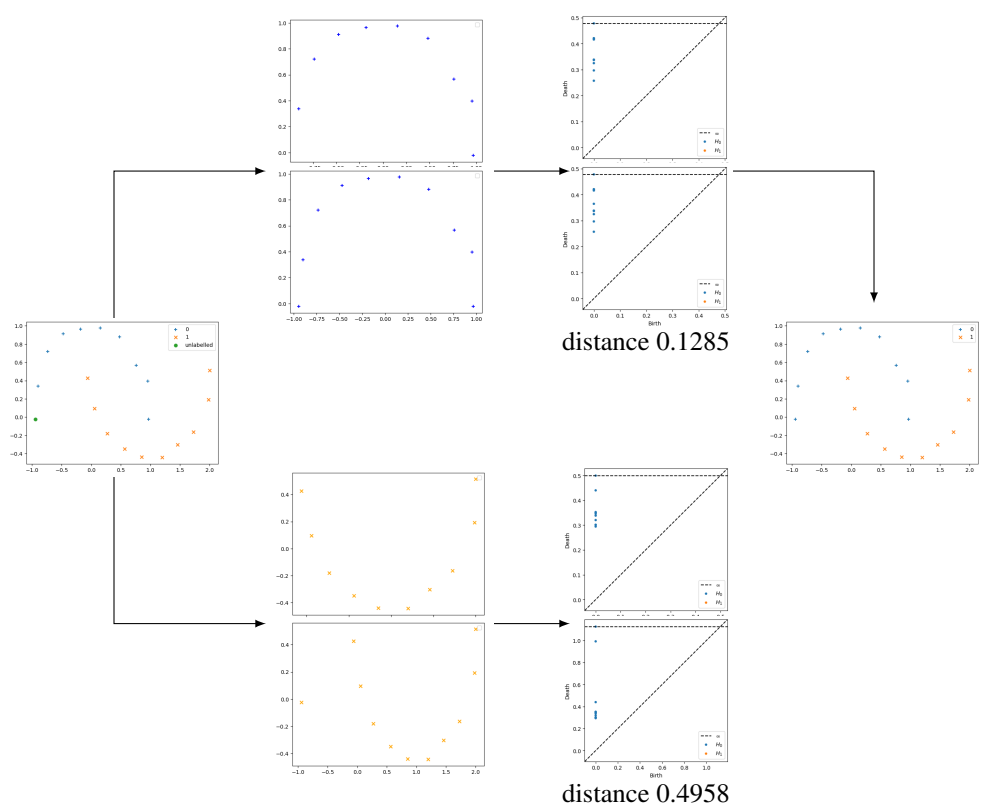
\begin{figure}[h]
\centering
\begin{tikzpicture}
\draw[-latex] (0,0) -- (0,3) -- (2,3);
\draw[-latex] (0,0) -- (0,-3) -- (2,-3);
\draw[-latex] (8,3) -- (10,3) -- (10,1.25);
\draw (0,0) node{\includegraphics[scale=0.2]{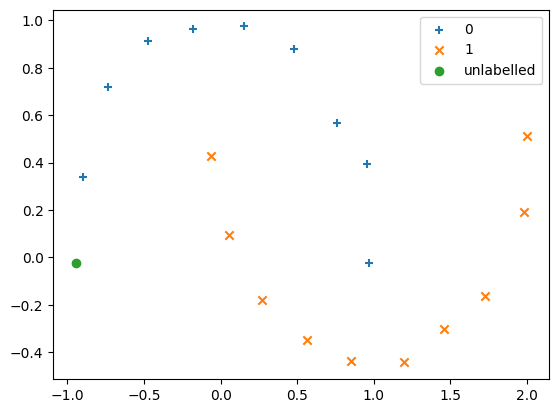}};

\draw (3.5,4) node{\includegraphics[scale=0.2]{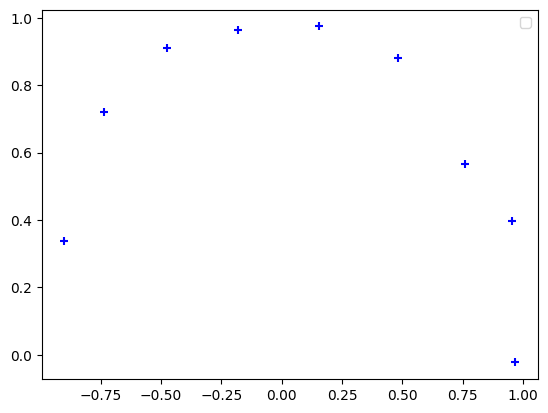}};
\draw (3.5,2) node{\includegraphics[scale=0.2]{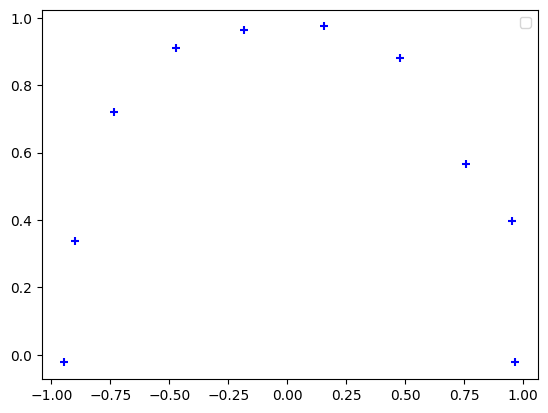}};
\draw (7,4) node{\includegraphics[scale=0.2]{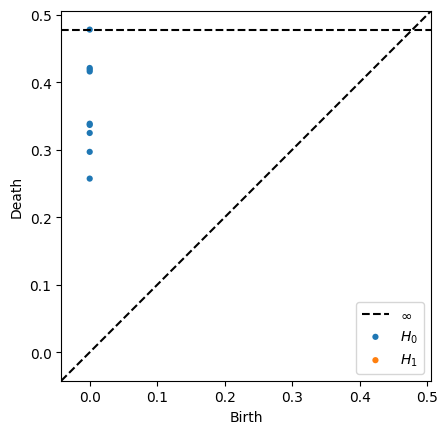}};
\draw (7,2) node{\includegraphics[scale=0.2]{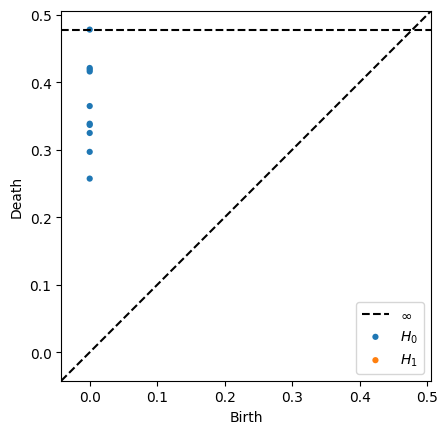}};
\draw[-latex] (5,3) -- (6,3);
\draw (7,0.75) node{{\small distance 0.1285}};

\draw (3.5,-2) node{\includegraphics[scale=0.2]{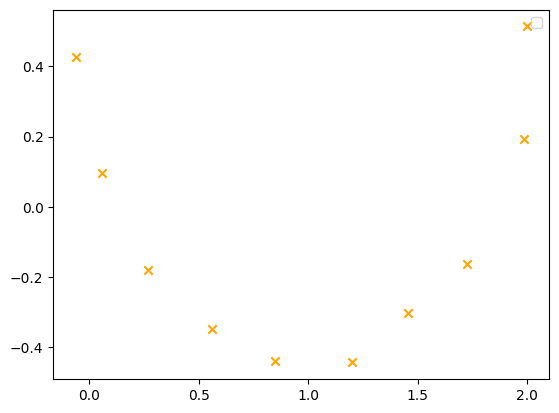}};
\draw (3.5,-4) node{\includegraphics[scale=0.2]{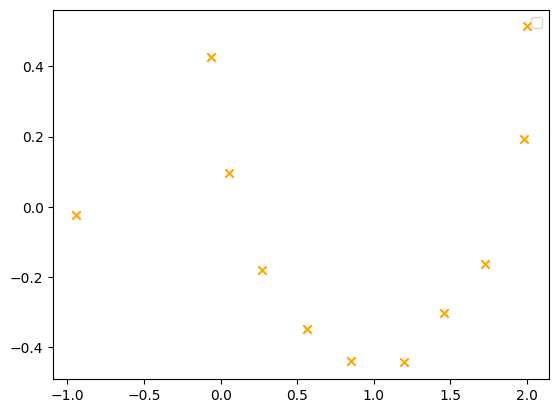}};
\draw (7,-2) node{\includegraphics[scale=0.2]{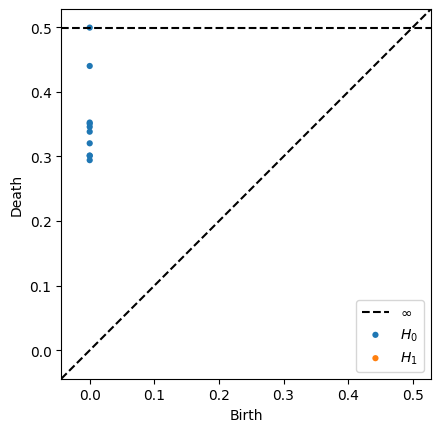}};
\draw (7,-4) node{\includegraphics[scale=0.2]{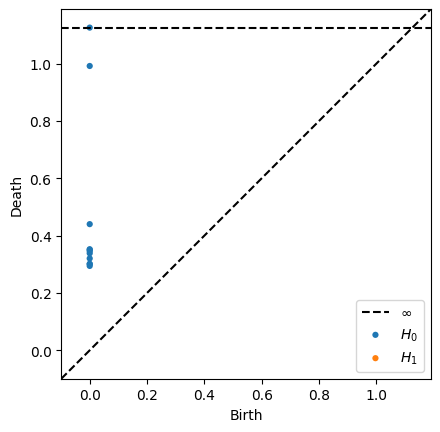}};
\draw[-latex] (5,-3) -- (6,-3);
\draw (7,-5.25) node{{\small distance 0.4958}};

\draw (10,0) node{\includegraphics[scale=0.2]{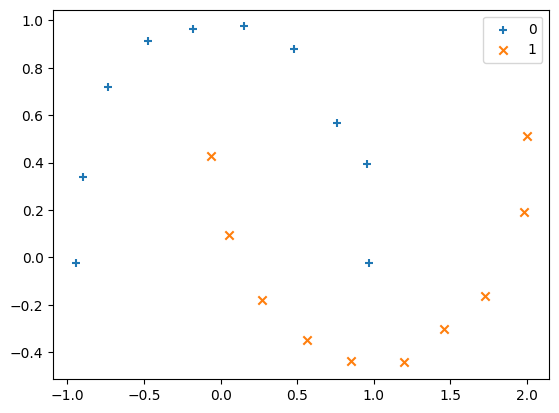}};

\end{tikzpicture}
    \caption{Example of the application of our method using the bottleneck distance, and using $0.6$ as threshold value.}
    \label{fig:exampleHomolo}
\end{figure}

The algorithm above is diagrammatically described in Figure~\ref{fig:exampleHomolo}, and it is applied to all the points of the set of unlabelled points $X$. 

\section{Evaluation}\label{sec:results}

Table~\ref{tab:structured} presents the results with 5 different datasets taken from the UCI Machine Learning Repository~\cite{datasets}, training the models with two machine learning algorithms, which are Support Vector Machines (SVM in the table) and Random Forest (RF), and comparing our method with three classical semi-supervised learning techniques (namely, Label Propagation~\cite{labelProp}, Label Spreading~\cite{labelSpread}, and Self Training) to annotate the unlabelled data. From these results, we can extract several conclusions: our method improves the base results in 8 out of the 10 models and obtains better results than the classical semi-supervised learning techniques in 8 out of the 10 models.

\begin{table*}
\caption{Accuracy results for the SVM and RF classifiers trained with data annotated for each of the annotation methods (classical and homological) together with the results obtained with the initial data (base) in the 5 structured datasets. The best result for each dataset is highlighted in bold face.}\label{tab:structured}
\centering
\resizebox{\linewidth}{!}{%
{\small
\begin{tabular}{ccccccccccccc}
 \toprule
   & \multicolumn{2}{c}{Banknote} & \multicolumn{2}{c}{Breast Cancer}  & \multicolumn{2}{c}{Ionosphere} & \multicolumn{2}{c}{Prima Indian} & \multicolumn{2}{c}{Sonar} & \multicolumn{2}{c}{Mean (std)}\\
  Method & SVM & RF & SVM & RF & SVM & RF & SVM & RF & SVM & RF & SVM & RF\\
 \midrule
 Base & 97.0 & 88.6 & 89.3 & \textbf{96.1} & 80.0 & 93.3 & 65.7 & 60.8 & 61.3 & 64.5 & 78.7(15.2) & 80.7(16.7)\\
 \midrule
 Label Propagation & 97.4 & 93.2 & 90.3 & 89.3 & 86.7 & 86.7 & 64.3 & 68.5 & 58.1 & 54.8 & 79.3(17.1) & 78.5(16.3)\\
 Label Spreading & 97.4 & 93.2 & 90.3 & 89.3 & 86.7 & 86.7 & 64.3 & 68.5 & 58.1 & 54.8 & 79.3(17.1) & 78.5(16.3)\\
 Self Training classifier & 95.1 & 93.6 & 35.9 & 35.9 & 85.0 & 86.7 & 66.4 & 66.4 & 58.1 & 67.7 & 68.1(23.2) & 70.1(22.4)\\
 \midrule
 Bottleneck threshold 0.8 & \textbf{99.2} & 92.4 & 93.2 & 91.3 & 78.3 & \textbf{95.0 }& 63.6 & 64.3 & 61.3 & 64.5 & 79.1(17.0) & \textbf{81.5(15.6)}\\
 Bottleneck threshold 0.6 & \textbf{99.2} & 91.3 & 89.3 & 90.3 & 75.0 & 88.3 & 59.4 & 63.6 & 48.4 & 45.2 & 74.3(20.9) & 75.7(20.6)\\
 Bottleneck threshold 0.4 & 97.4 & 90.5 & 87.4 & 85.4 & 78.3 & 86.7 & 63.6 & 62.9 & 45.2 & 45.2 & 74.4(20.5) & 74.1(19.5)\\
 Bottleneck threshold 0.2 & 97.4 & 90.5 & 87.4 & 85.4 & 78.3 & 86.7 & 63.6 & 62.9 & 45.2 & 45.2 & 74.4(20.5) & 74.1(19.5)\\
 Bottleneck threshold 0.0 & 97.4 & 90.5 & 87.4 & 85.4 & 78.3 & 86.7 & 63.6 & 62.9 & 45.2 & 45.2 & 77.1(22.6) & 74.1(19.5)\\
  \midrule
 Wasserstein threshold 0.8 & 97.4 & 89.8 & 92.2 & 88.4 & 80.0 & \textbf{95.0} & \textbf{68.5} & 67.8 & 61.3 & 64.5 & 79.9(15.3) & 81.1(13.9)\\
 Wasserstein threshold 0.6 & \textbf{99.2} & 93.6 & 89.3 & 87.4 & 70.0 & 91.7 & 61.5 & 61.5 & \textbf{74.2} & 61.3 & 78.9(15.2) & 79.1(16.3)\\
 Wasserstein threshold 0.4 & 97.0 & 96.2 & 87.4 & 87.4 & 76.7 & 81.7 & 60.8 & 62.9 & 71.0 & 71.0 & 78.6(14.1) & 79.8(13.2)\\
 Wasserstein threshold 0.2 & 97.0 & 96.2 & 87.4 & 87.4 & 76.7 & 81.7 & 60.8 & 62.9 & 71.0 & 71.0 & 78.6(14.1) & 79.8(13.2)\\
  Wasserstein threshold 0.0 & 97.0 & 96.2 & 87.4 & 87.4 & 76.7 & 81.7 & 60.8 & 62.9 & 71.0 & 71.0 & 78.6(14.1) & 79.8(13.2)\\
 \bottomrule
\end{tabular}}}
\end{table*}

\section{Conclusions and further work}

In this paper, we have studied the application of Topological Data Analysis techniques to the semi-supervised learning setting. The results show that our method can create classification models that achieve better results than those obtained when using classical semi-supervised learning methods. We plan to extend our work in different ways. First of all, the proposed method can be expanded to multi-class classification tasks, and, an iterative version of the algorithm can be easily developed. In addition, we plan to design new semi-supervised learning algorithms based on other notions from TDA, taking further advantage of the \emph{Manifold Hypothesis}. 



\end{document}